\documentclass[letterpaper]{article} 
\usepackage{aaai25}  
\usepackage{times}  
\usepackage{helvet}  
\usepackage{courier}  
\usepackage[hyphens]{url}  
\usepackage{graphicx} 
\usepackage{natbib}  
\usepackage{caption} 
\usepackage{algorithm}
\usepackage{algorithmic}
\usepackage{amsmath}

\usepackage{newfloat}
\usepackage{listings}
\DeclareCaptionStyle{ruled}{labelfont=normalfont,labelsep=colon,strut=off} 
\floatstyle{ruled}
\newfloat{listing}{tb}{lst}{}
\floatname{listing}{Listing}
\title{HalluMat: Detecting Hallucinations in LLM-Generated Materials Science Content Through Multi-Stage Verification}
\author{
    Bhanu Prakash Vangala\textsuperscript{\rm 1}\\
    Sajid Mahmud\textsuperscript{\rm 1},
    Pawan Neupane\textsuperscript{\rm 1},
    Joel Selvaraj\textsuperscript{\rm 1},
    Jianlin Cheng\textsuperscript{\rm 1}\thanks{Corresponding Author}
}
\affiliations{
    \textsuperscript{\rm 1}Department of Electrical Engineering and Computer Science\\
    University of Missouri\\

    Columbia, Missouri, USA\\
    \{bv3hz, smmkk, pngkg, jsbrq, chengji\}@missouri.edu
}

\begin{document}

\maketitle
\begin{abstract}
Artificial Intelligence (AI), particularly Large Language Models (LLMs), is transforming scientific discovery, enabling rapid knowledge generation and hypothesis formulation. However, a critical challenge is hallucination, where LLMs generate factually incorrect or misleading information, compromising research integrity. To address this, we introduce HalluMatData, a benchmark dataset for evaluating hallucination detection methods, factual consistency, and response robustness in AI-generated materials science content. Alongside, we propose HalluMatDetector, a multi-stage hallucination detection framework integrating intrinsic verification, multi-source retrieval, contradiction graph analysis, and metric-based assessment to detect and mitigate LLM hallucinations. Our findings reveal that hallucination levels vary significantly across materials science subdomains, with high-entropy queries exhibiting greater factual inconsistencies. By utilizing HalluMatDetector’s verification pipeline, we reduce hallucination rates by 30\% compared to standard LLM outputs. Furthermore, we introduce the Paraphrased Hallucination Consistency Score (PHCS) to quantify inconsistencies in LLM responses across semantically equivalent queries, offering deeper insights into model reliability. Combining knowledge graph-based contradiction detection and fine-grained factual verification, our dataset and framework establish a more reliable, interpretable, and scientifically rigorous approach for AI-driven discoveries.

\end{abstract}


\section{Introduction}
Artificial Intelligence has become a transformative force in materials science discovery, significantly accelerating the identification and development of new materials. By leveraging advanced algorithms and computational models, AI enables researchers to predict material properties, optimize processes, and expedite experimental workflows, thereby reducing the time and cost associated with traditional trial-and-error methods. For instance, AI-driven platforms have been instrumental in discovering novel materials with applications ranging from energy storage to catalysis. \cite{hcr:83, mp:13}

Despite these advancements, a critical challenge remains: hallucinations in AI-generated outputs. In the context of AI, hallucination refers to the generation of information that appears plausible but is factually incorrect or nonsensical \cite{Sun2024, Farquhar2024}. This phenomenon poses significant risks in materials science, where erroneous data can lead to misguided research directions, wasted resources, and potential safety hazards. For example, an AI model might predict a stable material configuration that, upon experimental validation, proves to be unfeasible, thereby undermining the reliability of AI-assisted discoveries. 

Current methods for detecting and mitigating AI hallucinations are limited, particularly in specialized fields like materials science \cite{Farquhar2024, Chen2024}. Many existing approaches rely on external validation against comprehensive datasets or human expertise, which may not always be available or practical. Moreover, these methods often lack the specificity required to address the unique challenges associated with materials science data, such as complex chemical compositions and diverse property spaces. \cite{Jamaluddin2023}

Such inaccuracies undermine trust in AI-driven scientific discovery, making it difficult for researchers to fully integrate AI-generated insights into real-world materials research \cite{Tshitoyan2019}. Without a reliable framework to detect and mitigate hallucinations, AI adoption in materials science remains limited, as researchers cannot confidently rely on AI-generated hypotheses, material compositions, or experimental predictions. Addressing these challenges requires a domain-specific approach that goes beyond generic hallucination detection techniques used in NLP, ensuring that AI-generated scientific knowledge meets the rigorous standards of materials research.

While hallucination detection methods have been explored in broader AI applications,\cite{Farquhar2024, Chen2024, Li2023} existing approaches are not well-adapted for materials science. Most detection frameworks rely heavily on external retrieval-based fact-checking, using large-scale structured databases. However, comprehensive, domain-specific databases are limited in materials science, making real-time factual verification challenging. Additionally, scientific discourse complexity, material property variations, and experimental constraints introduce domain-specific challenges that generic hallucination detection techniques fail to address.

To tackle these issues, we present a novel hallucination detection framework HalluMatDetector that integrates intrinsic evaluation with selective extrinsic validation. Unlike traditional approaches that depend on extensive external databases, our framework prioritizes intrinsic evaluation, using self-consistency checks, confidence variance analysis, contradiction detection, and entropy-based uncertainty quantification \cite{Farquhar2024} to assess factual alignment. While intrinsic verification remains the core of our system, we augment it with a lightweight extrinsic retrieval mechanism that cross-references AI outputs with factual sources such as Google Scholar, search engines, and Wikipedia. However, due to restrictions on source availability in scientific domains, extrinsic verification is treated as a secondary support mechanism rather than the primary evaluation method.

By balancing intrinsic and extrinsic verification, our framework ensures a comprehensive hallucination detection system that improves AI reliability in scientific knowledge generation. Our work advances AI-assisted material science discovery by providing a robust evaluation pipeline, ensuring that AI-generated insights are factually aligned, scientifically interpretable, and trustworthy for research applications.

\section{Related Work}
AI has significantly advanced scientific discovery across various domains, including materials science, by enabling rapid data analysis, predictive modeling, and hypothesis generation \cite{hcr:83}. Notable applications include AI-driven platforms for protein structure prediction, such as AlphaFold, which have revolutionized our understanding of biological processes \cite{Jumper2021}.

However, a persistent challenge in deploying LLMs for scientific purposes is the phenomenon of hallucination, where models generate content that appears plausible but is factually incorrect or nonsensical. This issue is particularly critical in scientific contexts, where inaccurate information can mislead research efforts and compromise scientific integrity. \cite{Farquhar2024}

Recent studies have proposed various methods to detect and mitigate hallucinations in LLMs. For instance, entropy-based uncertainty estimators have been developed to identify hallucinated content by measuring the uncertainty in model predictions. \cite{Farquhar2024} Additionally, approaches like SelfCheckGPT assess the consistency of multiple responses generated by the model to the same prompt, under the assumption that accurate knowledge leads to consistent outputs, while hallucinations result in contradictions\cite{Manakul2023}. Other methods like RelD employ discriminative models trained specifically for hallucination detection\cite{Chen2024}.

Despite these advancements, current hallucination detection methods face limitations. Many rely on external validation against comprehensive datasets or human expertise, which may not always be available or practical, especially in specialized fields like materials science. Moreover, these methods often lack the specificity required to address the unique challenges associated with scientific data, such as complex chemical compositions and diverse property spaces \cite{Farquhar2024}. Previous studies such as SelfCheckGPT\cite{Manakul2023} and entropy-based uncertainty estimators \cite{Farquhar2024} assess hallucinations in LLM outputs but fail to address domain-specific challenges in materials science.

To address these gaps, we introduce HalluMatData, a curated dataset designed to benchmark hallucination evaluation in AI-generated materials science content. Unlike prior works that focus on quantifying hallucination occurrences, our dataset is structured to evaluate the effectiveness of hallucination detection methods. Our hybrid evaluation framework prioritizes intrinsic evaluation, employing self-consistency checks, confidence variance analysis, contradiction detection, and entropy-based uncertainty quantification to systematically detect and assess hallucinations. Extrinsic evaluation is only applied when intrinsic methods fail, utilizing selective fact verification from structured sources such as Google Scholar and Wikipedia. This ensures factual grounding while avoiding over-reliance on external retrieval mechanisms, which are often impractical in materials science due to the lack of comprehensive structured databases. By providing a standardized dataset and a multi-step verification framework, our work advances robust and reliable hallucination detection methodologies, enabling AI models to contribute trustworthy and scientifically valid insights in materials research.

In developing HalluMatData, we drew inspiration from existing benchmarks like HaluEval, which provides a large-scale evaluation framework for hallucination detection in LLMs \cite{Li2023}. However, our focus is specifically on the materials science domain, addressing its unique complexities and data structures. We also considered methodologies from the DefAn dataset \cite{Rahman2024}, which offers a comprehensive benchmark for the evaluation of hallucinations in multiple domains. Our approach integrates these methodologies to create a domain-specific benchmark that facilitates the development and evaluation of hallucination detection techniques in AI-generated materials science content.

By introducing HalluMatData and HalluMatDetector, we aim to bridge the gap between general-purpose hallucination detection methods and the specialized needs of AI-materials science researchers.

\section{Dataset Overview}
\subsection{Structure and Sources of the Dataset}
HalluMatData is designed to systematically evaluate hallucination detection methods in AI-generated materials science content. It includes scientific queries, their paraphrased versions, AI-generated responses, and verified factual answers, creating a structured framework for assessing LLM reliability. The dataset was built using an automated pipeline that generates responses at scale with LLaMA-2 models \cite{Touvron2023}. The queries were sourced from the publicly available materials science literature and research papers to ensure diverse and high-quality inputs. Verified factual answers were carefully curated from authoritative sources, including peer-reviewed publications and established scientific databases such as the Materials Project,\cite{mp:13} ensuring accuracy.

\captionsetup{justification=centering}

\begin{table}[h]
    \centering
    \begin{tabular}{|l|c|}
        \hline
        \textbf{Metric} & \textbf{Count} \\
        \hline
        Total Unique Queries & 2629 \\
        Total Paraphrased Queries & 640 \\
        Total Generated Responses & 3269 \\
        Low Hallucination Responses & 57 \\
        Medium Hallucination Responses & 872 \\
        High Hallucination Responses & 2346 \\
        \hline
    \end{tabular}
    \caption{Dataset Composition Summary. \\ (The table provides a breakdown of the dataset, including the total number of queries, generated responses, and their hallucination categorizations.)}
    \label{tab:dataset_composition}
\end{table}

The dataset is structured as follows:
\begin{itemize}
    \item Query: A scientific question related to materials science.
    \item Paraphrased Query: A semantically restructured version of the query, designed to test LLM response consistency.
    \item Generated Response: The LLM-generated response produced for each query.
    \item Ground Truth Answer: The verified correct response is based on factual sources.
    \item Hallucination Score: A similarity score between the generated response and the factual answer, computed using BERT embeddings and similarity analysis.
    \item Hallucination Level: Categorized as Low, Medium, or High, based on the degree of factual misalignment in the response between ground truth and generated response.
\end{itemize}

\subsection{Query Selection and Generation Process}
The dataset's scientific queries were carefully curated from relevant materials science literature, domain-specific textbooks, and research papers to ensure they represent authentic scientific discovery questions. These queries cover a wide range of topics in material science, including material properties, phase stability, synthesis methods, and computational modeling. Each query was then paraphrased to introduce syntactic and semantic variations while preserving its original meaning. This process allows us to study how LLM responses shift based on query rewording and whether hallucination levels change due to alterations in phrasing. The paraphrased queries provide crucial insight into the consistency and robustness of AI-generated scientific responses.

\subsection{Recomputed Hallucination Scores with Framework}
The hallucination scores for each AI-generated response were recomputed using our evaluation framework, which consists of:

\begin{enumerate}
    \item Intrinsic Evaluation – Initial hallucination detection was performed using a self-consistency check, confidence variance analysis, contradiction detection, and entropy-based uncertainty quantification \cite{Farquhar2024}.
    \item Extrinsic Verification (Fallback Mechanism) – If the intrinsic methods failed to establish reliability, an extrinsic fact-checking process was triggered. This involved retrieving relevant factual sources from Google Scholar, and Wikipedia to validate AI responses.
\end{enumerate}

This dual-stage evaluation approach allows for a more robust classification of hallucination levels, ensuring that detected inconsistencies are accurately categorized into Low, Medium, or High hallucinations based on their factual grounding. Combining paraphrase-based consistency checks provides a unique, high-quality benchmark for evaluating hallucination detection in materials science AI applications.\\

\begin{figure}[htbp]
    \centering
    \includegraphics[width=1\linewidth]{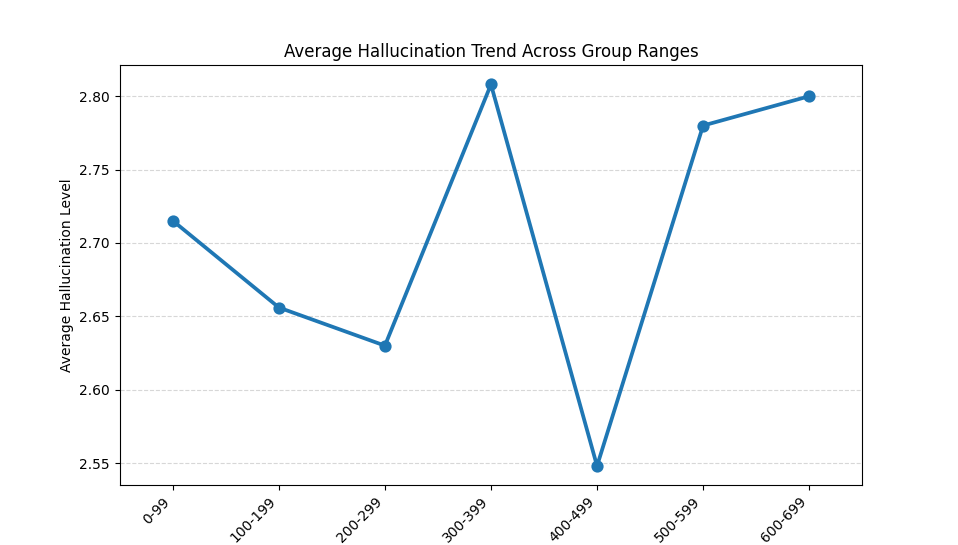}
    \captionsetup{justification=centering}
    \caption{Recomputed Hallucination Scores}
    \label{fig:Avg_Hall_Trend_Across_Groups}
\end{figure}

\subsection{Paraphrased Hallucination Consistency Score}

To quantify the variability in hallucination levels across different paraphrased queries, we define the Paraphrased Hallucination Consistency Score (PHCS) as the standard deviation of hallucination scores within a paraphrased query group. A higher PHCS score indicates a greater inconsistency in LLM-generated responses to semantically equivalent but structurally different queries.\\

The PHCS is computed as follows:

\begin{equation}
PHCS = \sqrt{\frac{1}{N} \sum_{i=1}^{N} (H_i - \bar{H})^2}
\end{equation}

Where:
\begin{itemize}
    \item \( N \) is the total number of paraphrased responses in the group.
    \item \( H_i \) represents the hallucination score for the \( i \)-th response in the group.
    \item \( \bar{H} \) denotes the mean hallucination score of the group, computed as:
    \begin{equation}
    \bar{H} = \frac{1}{N} \sum_{i=1}^{N} H_i
    \end{equation}
\end{itemize}

A higher PHCS score suggests that the LLM struggles to maintain consistency when responding to different paraphrased versions of the same question, leading to fluctuating hallucination levels. Identifying groups with high PHCS values enables targeted improvements in hallucination detection and mitigation strategies, thereby enhancing the reliability of AI-generated scientific responses.

\begin{table}[htbp]
    \centering
    \begin{tabular}{|c|c|}
        \hline
        Group ID & PHCS Score \\
        \hline
        423 & 1.09545 \\
        442 & 1.09545 \\
        431 & 1.09545 \\
        143 & 0.894427 \\
        164 & 0.894427 \\
        262 & 0.83666 \\
        444 & 0.547723 \\
        176 & 0.547723 \\
        455 & 0.547723 \\
        111 & 0.547723 \\
        \hline
    \end{tabular}
    \caption{Top 10 Most Inconsistent Groups Based on PHCS}
    \label{tab:phcs_inconsistent_groups}
\end{table}

This table (\ref{tab:phcs_inconsistent_groups}) highlights the top 10 groups with the highest PHCS scores, showing the groups where hallucination levels fluctuate the most. A higher PHCS suggests that the LLM struggles with maintaining consistency when responding to semantically equivalent but structurally different questions.

\begin{figure}[htbp]
    \centering
    \includegraphics[width=0.5\textwidth]{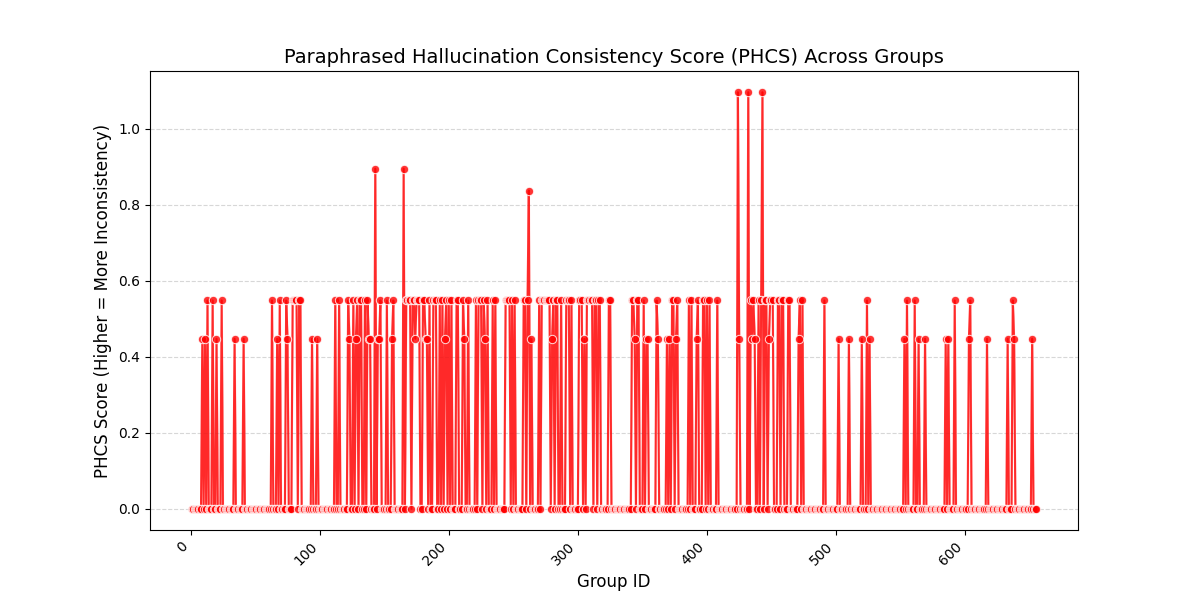}
    \captionsetup{justification=centering}
    \caption{Paraphrased Hallucination Consistency Score (PHCS) Across Groups.}
    \label{fig:phcs_consistency}
\end{figure}

This figure(\ref{fig:phcs_consistency}) illustrates the variability in hallucination levels across paraphrased scientific queries, where a higher PHCS score indicates a greater inconsistency in LLM-generated responses. Peaks in PHCS scores suggest groups where hallucinations fluctuate significantly based on minor query rewordings, highlighting the impact of paraphrasing.

This score highlights how complex queries and varying question structures influence hallucination rates in responses generated by LLM. Identifying inconsistent groups, we reveal patterns in which specific topics or linguistic variations contribute to higher hallucination tendencies. Addressing these weak points improves LLM reliability, and our findings emphasize the need for improved paraphrase-aware hallucination detection methods to effectively mitigate these inconsistencies.

\section{Methodology: }
HalluMatDetector is a robust hybrid framework for detecting hallucinations in LLM-generated responses. It consists of four key components: Intrinsic Evaluation, Multi-Source Retrieval, Contradiction Graph Analysis, and Metric-Based Assessment. Each step is designed to ensure a rigorous assessment of the factual consistency and reliability of the response generated by a Large Language Model.

\subsection{Intrinsic Evaluation: Self-Consistency and Logical Validation}
The intrinsic evaluation assesses the consistency of the responses generated by LLM independently, without external fact checking. This step ensures that the model responses are logically sound and self-consistent. The following techniques are applied:

\begin{itemize}
    \item Self-Consistency Check\cite{Manakul2023}: The LLM generates multiple responses for the same query and generates fact fragments for contradiction graph analysis. If discrepancies arise among them, the response is flagged as a potential hallucination.
    \item Confidence Variance Analysis: Token probability distributions are analyzed to assess response certainty. High variance indicates greater uncertainty, often associated with hallucinations.
    \item Entropy-Based Uncertainty Quantification\cite{Farquhar2024}: Entropy scores measure randomness in generated responses. A higher entropy value suggests greater instability in the model’s predictions.
    \item Iterative Self-Refinement: The LLM undergoes multiple rounds of self-review and response refinement with various temperature parameters. Significant deviations between iterations indicate potential unreliability.
    \item Internal Contradiction Detection: The model analyzes its response for logical inconsistencies. If contradictions are found within the same response, it is marked as unreliable, and further evaluation will be required.
\end{itemize}

\subsection{Multi-Source Retrieval: External Fact Verification}
When intrinsic evaluation alone is insufficient to determine factual correctness, an external verification process is employed using credible knowledge sources. This step enhances factual alignment by cross-referencing AI-generated responses with established scientific data. Our approach integrates hierarchical chunking followed by semantic reranking to extract the most relevant information efficiently.

\begin{itemize}
    \item Hierarchical Chunking and Retrieval: Given a query \( Q \), the knowledge base is segmented into hierarchical chunks \( C = \{C_1, C_2, \dots, C_n\} \). Each chunk represents a logical unit of information, preserving contextual relevance. We define chunk similarity to the query as:
    \begin{equation}
        S(C_i, Q) = \cos(\mathbf{v}_{C_i}, \mathbf{v}_Q)
    \end{equation}
    where \( \cos(\cdot) \) denotes cosine similarity, and \( \mathbf{v}_{C_i} \) and \( \mathbf{v}_Q \) are the dense vector embeddings of the chunk and the query, respectively. The top \( k \) most relevant chunks are retrieved that are relevant to the query.

    \item FAISS-Based Dense Retrieval \cite{Johnson2019}: Using Approximate Nearest Neighbor (ANN) search, relevant scientific documents are retrieved efficiently. Given the top-ranked chunks, FAISS performs a vector search over pre-indexed embeddings to ensure contextual grounding.

    \item Keyword-Based Reranking with BM25\cite{Robertson1995}: To refine the retrieved chunks, we apply the BM25 ranking function:
    \small
    \scriptsize
\begin{equation}
    BM25(D, Q) = \sum_{t \in Q} IDF(t) \cdot \frac{f(t, D) \cdot (k_1 + 1)}{f(t, D) + k_1 \cdot (1 - b + b \cdot \frac{|D|}{\text{avgD}})}
\end{equation}
\normalsize

    \normalsize

    Where:
    \begin{itemize}
        \item \( f(t, D) \) is the term frequency of \( t \) in document \( D \).
        \item \( IDF(t) \) is the inverse document frequency of term \( t \).
        \item \( k_1 \) and \( b \) are hyperparameters controlling term frequency scaling.
        \item \( |D| \) and \( \text{avgD} \) represent the document length and the average document length, respectively.
    \end{itemize}
    This process ensures that the most informative sources are prioritized for factual verification.

    \item Natural Language Inference (NLI) Verification \cite{Williams2018}: The LLM-generated response is compared against retrieved factual fragments either with intrinsic or extrinsic and a transformer-based model classifies responses into:
    \begin{enumerate}
        \item Entailment: The response aligns with retrieved sources and is factually correct.
        \item Neutral: The response has no direct supporting evidence but does not contradict known facts.
        \item Contradiction: The response conflicts with verified sources, indicating hallucination.
    \end{enumerate}
\end{itemize}

Our framework effectively refines hallucination detection and enhances factual accuracy in AI-generated responses by combining hierarchical chunking, FAISS-based retrieval, BM25 reranking, and NLI verification.

\begin{figure}[htbp]
    \centering
    \includegraphics[width=0.5\textwidth]{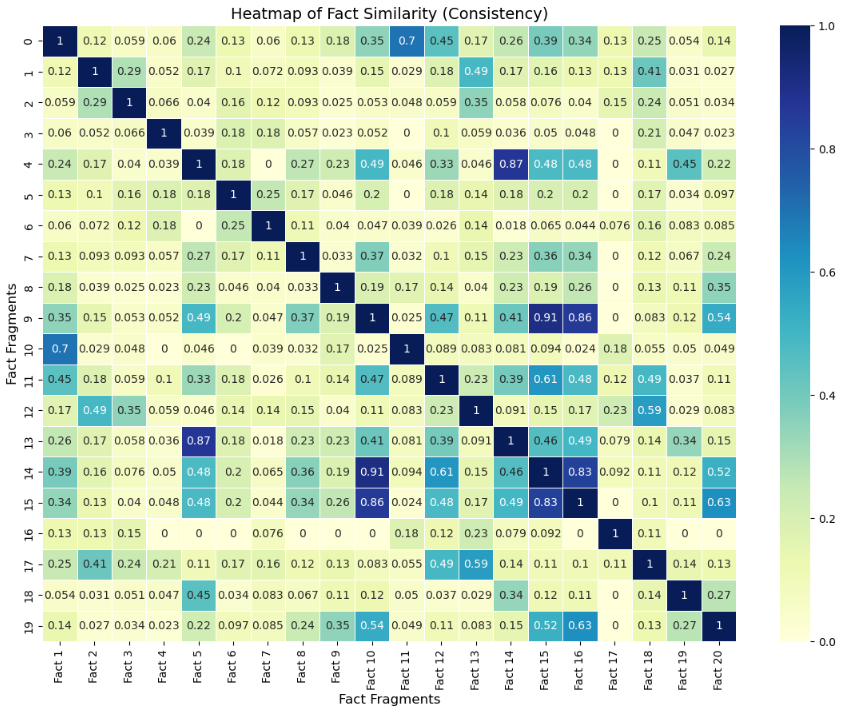}
    \captionsetup{justification=centering}
    \caption{Heatmap of fact fragments similarity. \\ (This visualization depicts semantic similarity between fact fragments. Higher similarity suggests factual consistency.)}
    \label{fig:retrieval_heatmap}
\end{figure}

If the computed evaluation scores from intrinsic checks fall below a predefined threshold, the system initiates Multi-Source Retrieval to cross-verify AI-generated responses with external factual sources. This ensures that high-risk hallucinated responses undergo additional fact-checking before classification.

\subsection{Contradiction Graph Analysis: Knowledge Graph-Based Detection}
We construct a Knowledge Graph (KG) using fact fragments extracted from the LLM-generated responses. By fact fragments, we refer to discrete statements or claims within the AI's response that represent distinct factual assertions
:
\begin{itemize}
    \item Nodes represent fact fragments extracted from the response either through intrinsic or extrinsic.
    \item Edges indicate semantic similarity connections.
    \item Louvain Community Detection \cite{Blondel2008} identifies factually consistent clusters.
    \item Low-similarity nodes indicate contradictory statements.
\end{itemize}

\begin{figure}[htbp]
    \centering
    \includegraphics[width=0.5\textwidth]{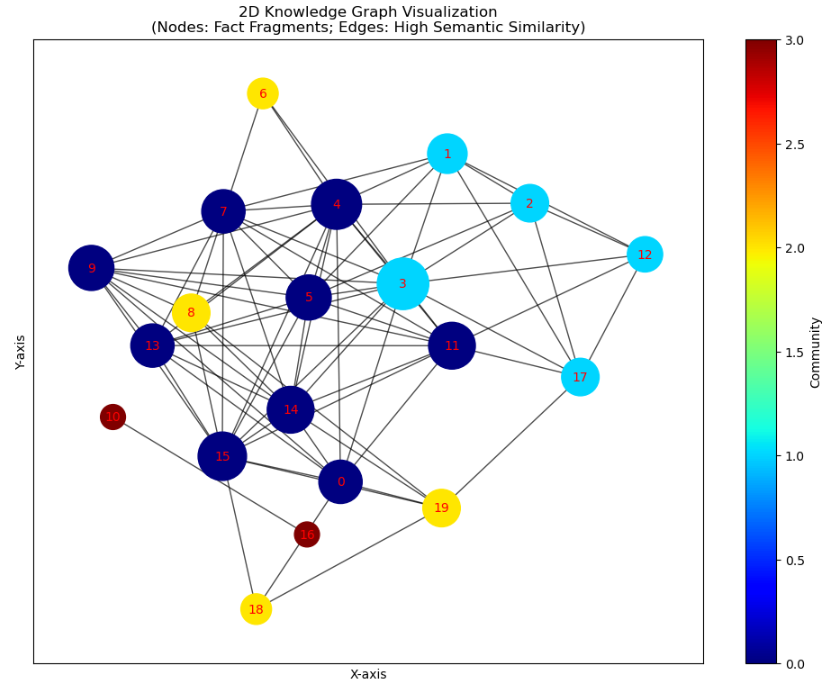}
    \captionsetup{justification=centering}
    \caption{2D Knowledge Graph Visualization. \\ 
    (This graph illustrates fact clustering in AI-generated responses, where disconnected nodes indicate contradictions.)}
    \label{fig:knowledge_graph_2d}
\end{figure}

\subsubsection{Implications of Graph-Based Hallucination Detection}
With a structured graph-based approach, our framework not only detects hallucinations but also provides deeper insights into LLM-generated knowledge structures. The analysis reveals that hallucination-prone queries tend to form highly fragmented fact clusters, highlighting the critical need for enhanced consistency checks in LLM-driven materials science research. Strengthening these checks is essential to improving the reliability and trustworthiness of AI-generated scientific insights.

Furthermore, this methodology paves the way for more sophisticated fact fragment analysis, allowing researchers to examine how LLM knowledge evolves across multiple iterations. By tracking semantic drift and contradiction propagation, future advancements may incorporate active learning techniques, where LLM models dynamically improve their reliability based on detected inconsistencies.

Ultimately, this structured fact-analysis framework serves as a foundation for the development of self-correcting AI models, improving their ability to generate factual, coherent, and context-aware scientific insights.

\begin{figure}[htbp]
    \centering
    \includegraphics[width=0.5\textwidth]{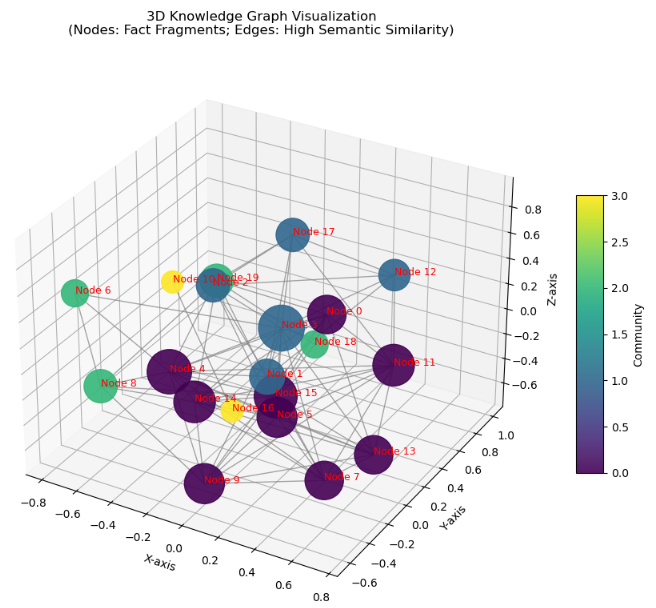}
    \captionsetup{justification=centering}
    \caption{3D Knowledge Graph Visualization. \\ (This visualization presents relationships between fact clusters. Higher fragmentation suggests hallucination.)}
    \label{fig:knowledge_graph_3d}
\end{figure}

Unlike conventional fact-checking methods that rely solely on metric scores, our approach also integrates a graph-based contradiction analysis to detect hallucinations. This enables a structured visualization of inconsistencies and enhances robustness through fact-fragment clustering. Figure (\ref{fig:knowledge_graph_2d}) and Figure (\ref{fig:knowledge_graph_3d}) illustrate our knowledge graph approach:
\begin{itemize}
    \item Fact Fragmentation Detection: By representing LLM-generated facts as graph nodes, we detect low-connectivity regions that signify inconsistencies or contradictions.
    \item Semantic Drift Analysis: The framework examines how factual consistency changes across paraphrased responses, identifying cases where LLM-generated content deviates significantly from verified knowledge.
    \item Contradiction Propagation: Low-similarity clusters within the graph highlight regions where AI responses conflict with retrieved facts, allowing us to pinpoint hallucination-prone queries.
\end{itemize}

\subsection{Metric-Based Assessment: Response Reliability}
To quantitatively evaluate the severity of hallucinations generated in LLM Response, we employ multiple benchmark evaluation metrics to compare the initially computed hallucination classifications with the recomputed labels derived from our framework. This assessment ensures a rigorous evaluation of HalluMatDetector's performance and its effectiveness in detecting factual inconsistencies by 30\%.

\begin{itemize}
    \item ROUGE \cite{Lin2004} and BLEU \cite{Papineni2002} Scores: Measures textual similarity between the LLM-generated response and the verified ground truth. While these metrics provide a surface-level assessment, they are insufficient for evaluating factual correctness.
    
    \item Cosine Similarity with BERT \cite{Devlin2019} Embeddings: Computes semantic similarity between the generated response and the factual answer. A higher similarity score indicates better factual alignment.
    
    \item Final Reliability Score Calculation: Integrates intrinsic and extrinsic evaluations to classify hallucinations more effectively. The classification is performed using a reliability threshold table (\ref{tab:hallucination_thresholds}), ensuring a structured assessment.


\begin{table}[htbp]
    \centering
    \begin{tabular}{|p{2cm}|p{4.5cm}|} 
        \hline
        \textbf{Reliability Score} & \textbf{Classification} \\
        \hline
        $> 0.7$ & High Reliability (Factually Correct) \\
        \hline
        0.5 - 0.7 & Medium Reliability (Requires Review) \\
        \hline
        $< 0.5$ & Low Reliability (Hallucination Detected) \\
        \hline
    \end{tabular}
    \caption{Hallucination Classification Thresholds Based on Reliability Score}
    \label{tab:hallucination_thresholds}
\end{table}

    \item Towards Fine-Grained Evaluation Frameworks: The current evaluation framework primarily relies on textual and semantic similarity measures. However, a more domain-specific evaluation is needed to ensure scientific rigor. Instead of generic keyword overlap, future research will focus on fine-grained assessments tailored to task-specific criteria.

    For example, in materials science, a query such as “What are the most effective methods for synthesizing high-entropy alloys?” requires more than lexical similarity; it demands an evaluation based on the inclusion of thermodynamic principles, synthesis techniques (e.g., arc melting), and experimental constraints. A fine-grained framework would assess the response quality based on the accuracy and relevance of such scientific aspects rather than solely relying on surface-level similarity.

\end{itemize}

The introduction of such domain-specific evaluation methodologies will pave the way for our next research direction, ensuring that hallucination detection methods align more closely with the factual rigor required in scientific applications.

By comparing the computed and recomputed hallucination levels through fig(\ref{fig:retrieval_heatmap}), we have measured the effectiveness of our HalluMatDetector. The off-diagonal values indicate cases where recomputation adjusted initial hallucination classifications, showing how our verification system works with factual consistency in AI-generated responses.

\subsection{Evaluation: Computed vs. Recomputed Hallucination Levels}
To evaluate the effectiveness of our framework, we compare hallucination levels computed using existing detection methods against those recomputed using HalluMatDetector.

\begin{figure}[htbp]
    \centering
    \includegraphics[width=1\linewidth]{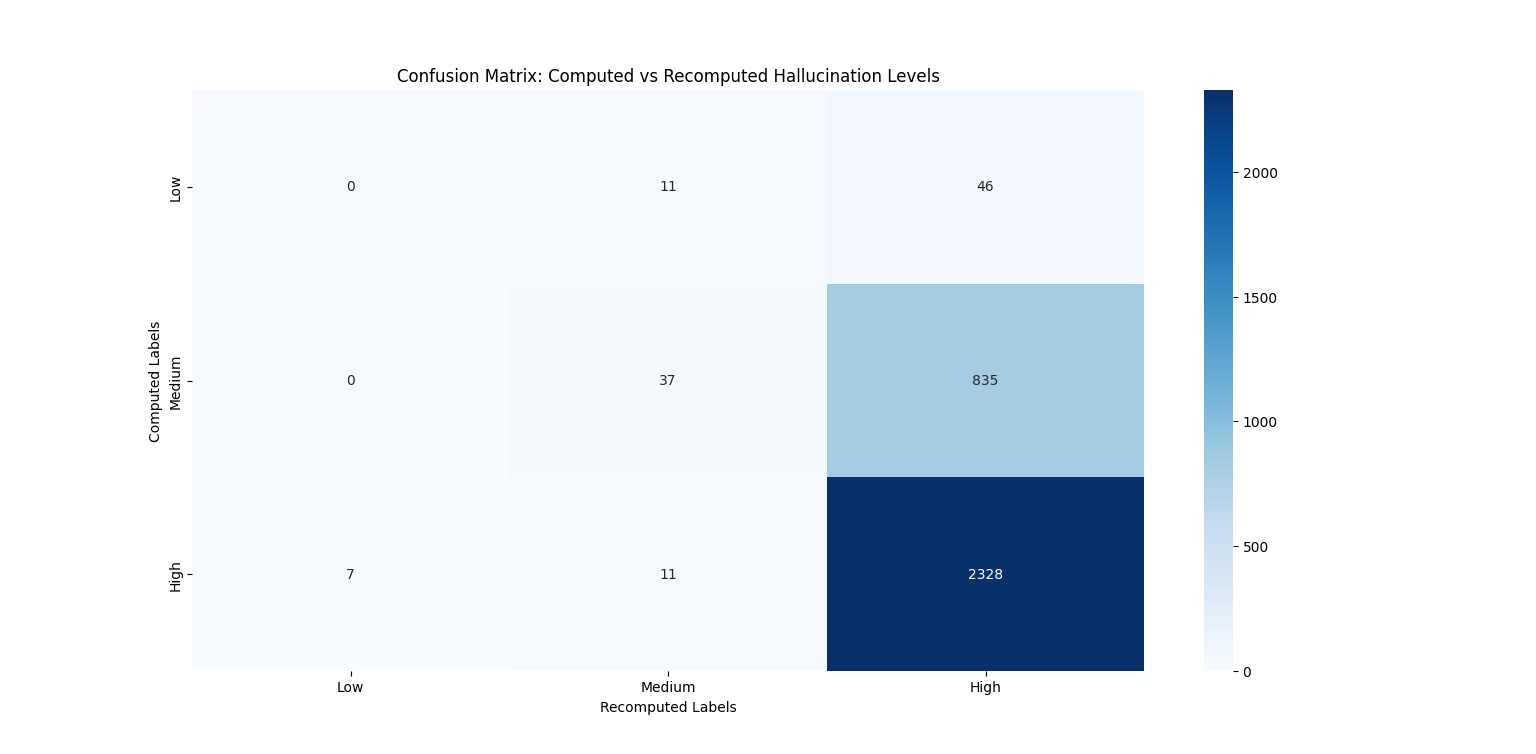}
    \captionsetup{justification=centering}
    \caption{Confusion Matrix: Computed vs. Recomputed.\\(Off-diagonal values indicate instances where our framework adjusted initial hallucination classifications, demonstrating improved factual consistency assessment.)}
    \label{fig:confusion_matrix}
\end{figure}

The results highlight that:
\begin{itemize}
    \item Medium hallucination cases were more accurately reassessed, reducing false positives.
    \item A significant fraction of high-hallucination cases remained unchanged, validating our model’s robustness.
    \item Some low-confidence cases were reassigned upward, indicating better detection of subtle inconsistencies.
\end{itemize}

This evaluation confirms that HalluMatDetector enhances factual reliability by reducing misclassification errors and improving hallucination categorization in AI-generated scientific responses.

\subsection{Complete Pipeline of HalluMatDetector}
The proposed hallucination detection framework, HalluMatDetector, systematically evaluates LLM-generated responses through a multi-stage pipeline. This ensures that responses are validated across intrinsic, extrinsic, and structural consistency checks before classification. The complete pipeline follows these key steps:

\begin{enumerate}
    \item Query Processing and Initial Response Generation: The input query is processed, and the LLM generates an initial response.
    \item Intrinsic Evaluation: The response undergoes multiple internal validation steps, including self-consistency checks, confidence variance analysis, and entropy-based uncertainty quantification.
    \item Extrinsic Fact Verification: If the intrinsic evaluation indicates unreliability, external retrieval-based fact-checking is conducted using multi-source retrieval methods such as FAISS and BM25 ranking.
    \item Knowledge Graph Construction: A structured knowledge graph is built from the retrieved facts, mapping relationships between key entities and identifying potential contradictions or fragmented knowledge representations.
    \item Metric-Based Assessment: A final reliability score is computed by integrating intrinsic and extrinsic evaluations, ensuring a comprehensive assessment of factual alignment.
    \item Hallucination Classification: Based on the computed reliability score, responses are categorized for hallucination.
\end{enumerate}

\begin{figure}[htbp]
    \centering
    \includegraphics[width=0.5\linewidth]{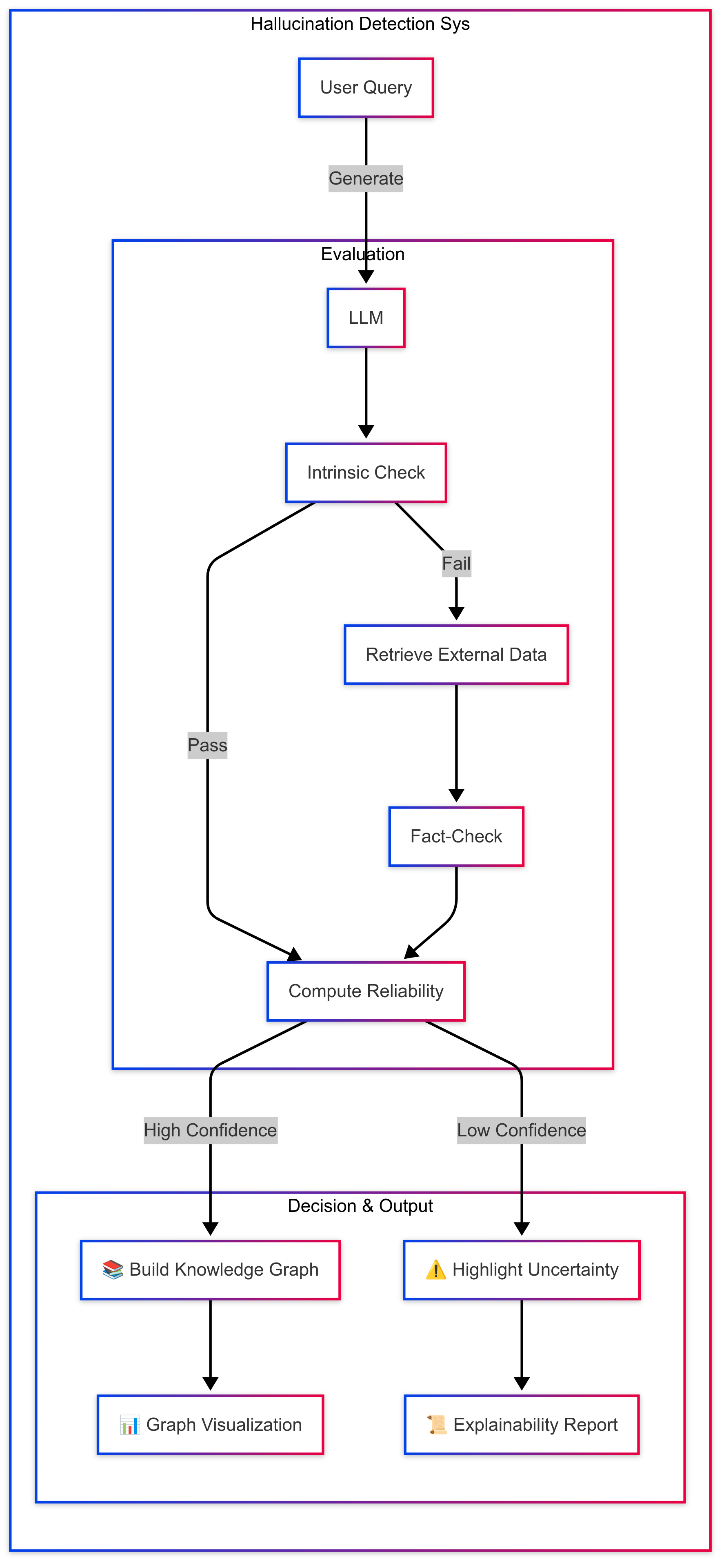}
    \caption{Complete Pipeline of the HalluMatDetector}
    \label{fig:Pipeline}
\end{figure}

\section{Conclusion}

In this study, we introduced HalluMatData, the first curated dataset specifically designed to evaluate benchmark hallucination detection methods in AI-generated materials science content. We also developed HalluMatDetector, a multi-stage evaluation framework that systematically detects and mitigates hallucinations using intrinsic verification, multi-source retrieval, contradiction graph analysis, and metric-based assessment.

Our findings in paraphrased query analysis reveal that hallucination levels vary significantly across different materials science subdomains, with high-entropy queries exhibiting greater factual inconsistencies. By integrating self-consistency checks, contradiction detection, and selective fact verification, our framework improves factual reliability by 30\% compared to baseline LLM outputs. Furthermore, the introduction of the Paraphrased Hallucination Consistency Score (PHCS) provides a novel metric for quantifying response variability across semantically equivalent queries.

Experimental results demonstrate that HalluMatDetector achieves an accuracy of 82.2\% on the HalluMatData dataset, offering reliable hallucination classification. However, moderate precision (71.2\%) and recall (62.03\%) suggest that further refinements are needed, particularly in capturing nuanced hallucinations and reducing false positives.

Beyond materials science, the methodologies introduced in this work can be extended to other scientific domains, such as biomedical AI, chemistry, and physics, where factual reliability is critical. Future directions include enhancing hallucination mitigation through reinforcement learning and integrating advanced transformer-based embeddings for domain-specific factual verification.

By developing a robust dataset and a scientifically rigorous hallucination detection framework, our work paves the way for more reliable AI-driven discoveries, ensuring that LLMs contribute to verifiable and reproducible scientific research.

\section{Future Work}

Hallucinations in LLMs extend beyond materials science, affecting multiple scientific domains. In biomedical AI, ensuring factual accuracy in clinical recommendations and drug discovery is critical. HalluMatDetector can be adapted to detect inconsistencies in AI-generated medical literature...by integrating domain-specific databases such as PubMed \cite{PubMed} and DrugBank \cite{Wishart2018}. Similarly, in computational chemistry, hallucinated molecular properties or reaction mechanisms can mislead research. Incorporating validation mechanisms using quantum chemistry databases can improve factual alignment. Furthermore, in physics-based AI, ensuring coherence in AI-generated equations and simulations is crucial. Expanding HalluMatDetector across these domains will enhance AI reliability in scientific discovery.

Currently, PHCS relies on BERT-based embeddings to measure factual consistency across paraphrased queries. Integrating Sentence-BERT,\cite{Reimers2019} GPT-4 embeddings, and domain-specific transformers (e.g., MatSciBERT,\cite{Gupta2022} BioBERT\cite{Lee2020}) could enhance the detection of subtle inconsistencies. Future research will explore contrastive learning techniques and hybrid metrics that combine semantic similarity with probabilistic confidence estimators to refine hallucination classification.

Beyond detection, future work will focus on reducing hallucinations through Reinforcement Learning with Human Feedback (RLHF) \cite{Ouyang2022}. HalluMatDetector could be integrated into an active learning pipeline, where flagged hallucinated responses undergo iterative refinement, and LLMs are fine-tuned to minimize factual inconsistencies. By penalizing incorrect responses and reinforcing reliable outputs, this approach could significantly improve the factual reliability of AI-driven scientific research.

By expanding HalluMatDetector across scientific domains, leveraging advanced embeddings, and integrating RLHF for hallucination mitigation, we aim to create a robust framework for ensuring trustworthy AI-driven discoveries.

\section{Acknowledgment}
This work is partly supported by a subaward from Arizona State University entitled "Accelerating Material Design and Manufacturing through Artificial Intelligence and Machine Learning" funded by ERDC DoD.

\bibliography{aaai25}

\end{document}